# Role-Playing Simulation Games using ChatGPT

by Rita Stampfl (University of Applied Sciences Burgenland), Igor Ivkić (University of Applied Sciences Burgenland and Lancaster University) and Barbara Geyer (University of Applied Sciences Burgenland)

*Since the COVID-19 pandemic, educational institutions have embarked on digital transformation projects. The success of these projects depends on integrating new technologies and understanding the needs of digitally literate students. The "learning by doing" approach suggests that real success in learning new skills is achieved when students can try out and practise these skills. In this article, we demonstrate how Large Language Models (LLMs) can enhance the quality of teaching by using ChatGPT in a role-playing simulation game scenario to promote active learning. Moreover, we discuss how LLMs can boost students' interest in learning by allowing them to practice real-life scenarios using ChatGPT.*

The COVID-19 pandemic challenged educators to create engaging and innovative distance learning environments. Instead of relying solely on lectures and traditional homework, teachers had to use digital tools to enrich student participation in online sessions. Moreover, to maintain high levels of motivation and interest, it became necessary to be more inventive with assignments given between classes. This emphasis on creating unique and engaging assignments has persisted beyond the pandemic period.

Meanwhile, LLMs, especially with the introduction of chatbots such as ChatGPT that use artificial intelligence (AI) to communicate with users, have received considerable attention in higher education. When used responsibly, LLMs can enhance distance learning tasks [1] and contribute to the development of metacognitive skills such as critical thinking and problem solving [L1]. As higher education institutions strive to prepare students for the evolving labour market, the use of LLM tools has become increasingly important within the Future Skills Framework [L2].

According to Statistics Austria, 96.33% of students enrolled at Austrian Universities of Applied Sciences in the winter semester of 2022/23 were under 40 years old [L3]. These students, who were born after 1980, grew up in the digital world and were influenced by computer technology. They developed through various forms of media, including videos, consoles and games. Such games capture players' attention through enjoyable and entertaining activities. By immersing students in real-world situations, game learning uses theories and applications that students must use to improve learning engagement and performance [2]. Simulation games are a unique and effective teaching method that increases students' motivation to learn [3]. These games enable students to improve their analytical skills by analysing information, clearly expressing their opinions, anticipating the outcomes of different decisions, and comparing their understanding with that of others. Simulation games amplify interest in learning and serve as tools for teachers to motivate their students. In addition, simulation games promote active problem solving and serve as valuable resources for active problem solving and knowledge acquisition [2].

In a similar study, Schmid et al. [L4] used LLM tools for multimedia human-machine interaction, and Matute Vallejo and Melero [2] considered simulation games as tools for active problem solving and immersion in real situations. Based on their research, we propose a method for using an LLM tool in a simulation game allowing students to apply their learning in practical scenarios. Specifically, ChatGPT can be used as a chatbot for simulating conversational scenarios. This allowed the students to engage with the game independently, overcoming time and location constraints. The advantage is that ChatGPT plays the role of an interlocutor in asynchronous gameplay.

To implement this concept, we conducted a case study for the course "Impact of Cloud Computing on Organisations" from the master's programme "Cloud Computing Engineering" at the University of Applied Sciences Burgenland [L5]. The main objective of the course is to educate students about the impact of cloud technologies on international, multicultural organisations and to develop their social skills in areas such as change management, negotiation and decision-making. To assess these skills, an assignment was created in which students participated in a role-playing negotiation meeting using ChatGPT. The students were also required to create a proposal for a cloud migration project during the course. The simulation for this project assignment was based on the benefits of cloud computing as identified by Salesforce [L6]. As shown in Figure 1, a prompt was created and given to the students which initiated the role-play activity in ChatGPT. During the role-playing game, the students were given the role of advocating and negotiating for the approval of the project budget.

The conversation in ChatGPT was initiated with a fixed prompt (Figure 1), which allowed for different flows and out-

> Let's role play. You are the CEO of a company that is thinking about moving all its on-site hosted servers, including its services, to the cloud. However, you are not sure about this decision and are very critical of the cloud in general. I want to enter a project budget of >>amount from your project order<< Euro and you want to give it to me only after intense negotiations. You ask me critical questions about the following areas: Cost savings, security, flexibility, mobility, insight, increased collaboration, quality control, disaster recovery, loss prevention, automatic software updates, competitive advantage and sustainability. I am a cloud consultant who answers your questions. You ask one question at a time and ask the next question based on my answer. I only get a commitment from you if the IPMA project assignment criteria are met.

*Figure 1: Prompt for starting a role-playing simulation game using ChatGPT.*

comes for each student. However, this scenario mirrors real-life negotiations with organisational management, where students need to effectively present compelling arguments to secure budget approval. In addition to role-playing, students should also reflect on their personal experiences during the simulation. These reflections can inform future assignments and aid the development of new simulation games.

The presented idea demonstrates that LLM tools can be used for more than just generating written output based on a given input. Specifically, with a prompt, as shown in Figure 1, ChatGPT can be programmed to initiate a conversation with its user, thereby simulating a real-life scenario for practice. This approach was first tested in the master's programme, as described in the case study above. It is worth noting that this example serves as an initial exploration of the integration of LLM into role-playing simulations in the classroom. Although not comprehensive or conclusive, this study intends to inspire educators to rethink their course designs and develop new interactive tasks using similar methods. In future research, we will extend our approach and test different prompts in different scenarios and master's programmes. We plan to conduct an in-depth case study involving focus groups from four master's programmes and create different role-plays for each of them. We aim to evolve our methodology further and assess its broader applicability across a more expansive spectrum of applications.

**Links:**
[L1] https://kwz.me/hAk
[L2] https://www.stifterverband.org/future-skills/framework
[L3] https://kwz.me/hAn
[L4] https://kwz.me/hAN
[L5] https://kwz.me/hAe
[L6] https://kwz.me/hAP

**Please contact:**
Rita Stampfl, University of Applied Sciences Burgenland, Austria
rita.stampfl@fh-burgenland.at

# ChatGPT as a Learning Assistant in Distance Learning

by Michael Prodinger, Rita Stampfl and Marie Deissl-O'Meara (University of Applied Sciences Burgenland)

*The following article deals with the implementation of a Learning Assistant, an advanced tool based on artificial intelligence (AI) that provides continuous learning support to students. The assistant is used in distance learning programmes at FH Burgenland Continuing Education. With ChatGPT installed and integrated into the Learning Management System (LMS), it functions as an assistant with the ability to answer questions from students whenever they arise. Additionally, a course teacher verifies and corrects the AI's responses within 24 hours to guarantee the system's correctness and dependability.*

The ongoing digitalisation and transformation of higher education (HE) creates new opportunities for innovative teaching methods, especially in the area of distance learning. Teachers should thus concentrate on how digital tools are transforming education. It is therefore crucial that they develop strategies to offer attractive digital education options to learners [1]. The proposed addition of a Learning Assistant to the FH Burgenland Continuing Education distance learning programmes [L1] is one specific illustration of this development. A closer examination of the several facets of this innovation, how it works, and the associated added value for students, is conducted in light of its successful implementation at other institutions [L2].

The Learning Assistant is designed as an AI tool that provides interactive learning support to students at any time of the day or night. Its main function is to answer students' content-related questions, using only the content stored in the LMS. This specific configuration ensures that the information and answers provided are always relevant to the programme and aligned with the curriculum. The integration of the Learning Assistant into the LMS therefore represents a targeted extension of the existing resources, with the aim of providing students with an additional information channel for acquiring knowledge and clarifying any uncertainties. It is clear that learning assistants will permanently change the way we teach and learn [L3].

At the heart of the Learning Assistant is an AI system based on the Large Language Model (LMM). This is an advanced technology capable of generating and understanding human-like text [L4]. LLMs have gained prominence in HE, especially with the integration of ChatGPT, which uses AI to interact with users. If used responsibly by all stakeholders, LLMs can improve the quality of distance learning assignments [2]. In the context of distance learning, the LLM is configured to use the content stored in the LMS as the basis for its responses.

A key benefit of the Learning Assistant is its availability. Students may engage in conversation with the AI and pose questions whenever necessary. This is especially crucial for